%% file: main_arxiv.tex
\documentclass[10pt,journal,compsoc]{IEEEtran}

\ifCLASSOPTIONcompsoc
  \usepackage[nocompress]{cite}
\else
  \usepackage{cite}
\fi
\ifCLASSINFOpdf
\else
\fi

\usepackage{url}
\usepackage{amsmath,amsfonts}
\usepackage{algorithm}
\usepackage{array}
\usepackage[caption=false,font=normalsize,labelfont=sf,textfont=sf]{subfig}
\usepackage{textcomp}
\usepackage{stfloats}
\usepackage{url}
\usepackage{verbatim}
\usepackage{graphicx}
\usepackage{cite}
\usepackage{times}
\usepackage{epsfig}
\usepackage{graphicx}
\usepackage{amsmath}
\usepackage{amssymb}
\usepackage{booktabs}
\usepackage{bm}         
\usepackage{amsmath}         
\usepackage{graphicx}
\usepackage{multirow, tabularx}
\usepackage{makecell}
\usepackage{pifont}
\usepackage{wrapfig}
\usepackage{caption}
\usepackage{graphicx}
\usepackage{comment}
\usepackage{xspace}
\usepackage{textcomp}%
\usepackage{manyfoot}%
\usepackage{algorithm}%
\usepackage{algpseudocode}%
\usepackage{listings}%
\usepackage{ragged2e}
\usepackage{xcolor}

\newcommand{\ie}{\emph{i.e.}\xspace}
\newcommand{\eg}{\emph{e.g.}\xspace}

\newcommand{\nickname}{HumanLiff}

\begin{document}
%

\title{HumanLiff: Layer-wise 3D Human Generation with Diffusion Model}

%
%
%
%

\author{Shoukang Hu,
Fangzhou Hong,
Tao Hu,
Liang Pan,
Haiyi Mei,
Weiye Xiao,
Lei Yang,
Ziwei Liu
\IEEEcompsocitemizethanks{
\IEEEcompsocthanksitem Shoukang Hu, Fangzhou Hong, Tao Hu, Liang Pan and Ziwei Liu are with S-Lab, Nanyang Technological University, Singapore. 
\IEEEcompsocthanksitem Haiyi Mei, Weiye Xiao and Lei Yang are with SenseTime, China. 
}
}

\input{latex/0_abstract.tex}

\maketitle

\IEEEdisplaynontitleabstractindextext

%
\IEEEpeerreviewmaketitle


\input{latex/1_introduction.tex}
\input{latex/2_related_work.tex}

\input{latex/3_method.tex}
\input{latex/4_experiment.tex}
\input{latex/5_discussion.tex}
\ifCLASSOPTIONcaptionsoff
  \newpage
\fi



{
\bibliographystyle{IEEEtran}
\bibliography{IEEEabrv,egbib}
}

\end{document}

%% file: latex/0_abstract.tex
\IEEEtitleabstractindextext{
\begin{abstract}

3D human generation from 2D images has achieved remarkable progress through the synergistic utilization of neural rendering and generative models.
Existing 3D human generative models mainly generate a clothed 3D human as an undetectable 3D model in a single pass, while rarely considering the layer-wise nature of a clothed human body, which often consists of the human body and various clothes such as underwear, outerwear, trousers, shoes, etc.
In this work, we propose \textbf{\nickname}, the first layer-wise 3D human generative model with a unified diffusion process.
Specifically, \nickname{} firstly generates minimal-clothed humans, represented by tri-plane features, in a canonical space, and then progressively generates clothes in a layer-wise manner.
In this way, the 3D human generation is thus formulated as a sequence of diffusion-based 3D conditional generation. 
To reconstruct more fine-grained 3D humans with tri-plane representation, we propose a tri-plane shift operation that splits each tri-plane into three sub-planes and shifts these sub-planes to enable feature grid subdivision.
To further enhance the controllability of 3D generation with 3D layered conditions, \nickname\ hierarchically fuses tri-plane features and 3D layered conditions to facilitate the 3D diffusion model learning.
Extensive experiments on two layer-wise 3D human datasets, SynBody (synthetic) and TightCap (real-world), validate that \nickname \ significantly outperforms state-of-the-art methods in layer-wise 3D human generation.
Our code will be available at \url{https://skhu101.github.io/HumanLiff}.


\end{abstract}

}

%% file: latex/1_introduction.tex
\section{Introduction}

\begin{figure*}[t]
    \centering
    \includegraphics[width=1.0\linewidth]{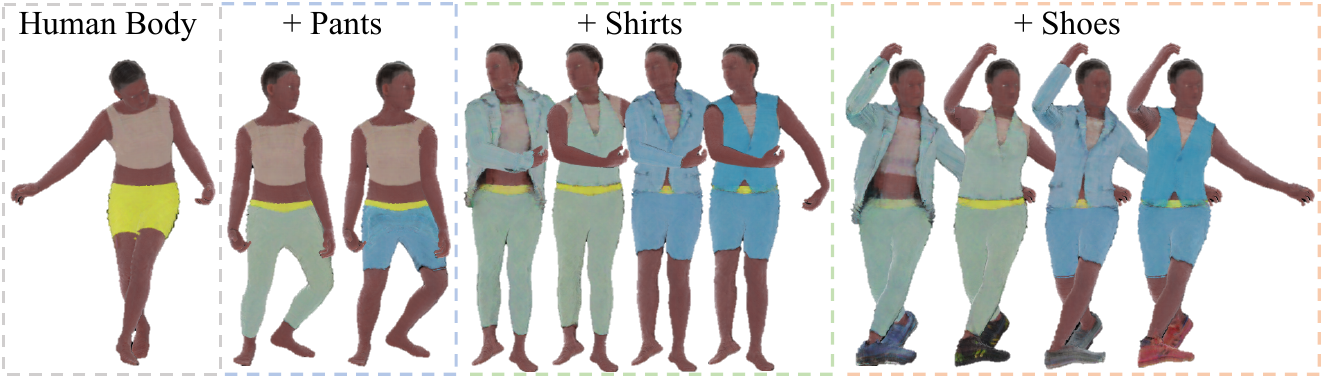}
    \setlength{\abovecaptionskip}{-0.3cm}
    \caption{\nickname{} learns to generate layer-wise 3D human with a unified diffusion process. Starting from random noise, \nickname{} first generates a human body and then progressively generates 3D humans conditioned on the previous generation. We use the same background color to denote generation results from the same human layer.} 
\label{fig: teaser}
\vspace{-5mm}
\end{figure*}

\IEEEPARstart{G}{enerating} 3D clothed humans with various clothing styles are required in many applications such as AR/VR, telepresence, game and film character creation. 
To create 3D clothed humans, existing approaches either require heavy artist work or a large dataset of 4D scans to learn human reconstructions \cite{Saito2021SCANimateWS}. 
However, these solutions are expensive, and instead, learning a data-driven model to generate 3D clothed humans from scalable 2D image collections will automate the creation and enable many applications for 3D content generation.

Most existing data-driven approaches for human generation produce human representations from pose- and view-conditioned noises, which can further be rendered by 2D  \cite{Fruhstuck2022InsetGANFF,Fu2022StyleGANHumanAD} or 3D GAN-based renderers \cite{noguchi2022unsupervised, bergman2022generative, hong2022eva3d}. 
However, these methods generate human images in a one-stage noise-to-human manner assuming the garment and body skin belong to the same surface layer and rarely consider the layer-wise nature of 3D Humans, \ie, the final 3D clothed human is a layer-wise composition of human bodies, pants, shirts and shoes \textit{etc.}
Thus these methods lose control over the whole generation process, which poses challenges to cases where users hope to generate several choices in each layer generation.
For example, in VR/AR, game players may hope to create game characters layer by layer, generating a minimal-clothed human body and then generating/selecting pants, shirts, and shoes. 

In this work, we propose \textbf{\nickname}, the first layer-wise 3D clothed human generation approach using diffusion model, which allows users to freely control the generation of human bodies, pants, shirts, and shoes, \textit{etc} (as shown in Fig.~\ref{fig: teaser}). 
Specifically, \nickname\ encodes layer-wise 3D humans with tri-plane representations in a canonical space and learns to produce the 3D human representations in a layer-wise manner (as shown in Fig.~\ref{fig: overview}).
We formulate the layer-wise generation task as a sequence of diffusion-based conditional generation, where the generations of each garment layer are conditioned on previously dressed layers.
There are three critical aspects in this sequential generation task: \textbf{1)} how to model 3D humans; \textbf{2)} how to dress 3D humans with different garment layers; and \textbf{3)} how to control the layer-wise 3D human generation.

The first aspect lies in the choice of 3D human representation. Most recent approaches encode 3D scenes in fully connected layers (e.g., NeRF \cite{mildenhall2020nerf}) or explicit voxel grids~\cite{sun2022direct, fridovich2022plenoxels}. 
The former is slow to query, and though voxel grids converge fast in training, they scale poorly with resolution since the high space complexity of voxel grids limits the training resolution ~\cite{muller2022diffrf}. 
In contrast, we employ the tri-plane representation \cite{chan2022efficient} and extend it to model articulated human bodies, which are fast to query and capable of producing high-fidelity 3D humans.
To reconstruct more fine-grained 3D humans, we equip tri-plane representation with \textbf{tri-plane shift} (as shown in Fig.~\ref{fig: tri-plane shift}), which splits each tri-plane into three sub-planes in the channel manner and shift one sub-plane half tri-plane unit to the right, and one sub-plane half tri-plane unit downward.
With this design, 3D sampled points projected to the same square area in each tri-plane will not hold the same features. Instead, extracted features can be used for rendering more fine-grained results without enlarging the total number of parameters.

The second aspect is how different garment layers are dressed on 3D human bodies, and a key component is to align garments with body shapes. 
To tackle this challenge, garment styles are decoupled into different layers (e.g., pants, skirts, shoes) according to the 3D human generation process in the current graphics pipeline, and we model both humans and garments in a canonical space, where the body shapes and garments are aligned based on the correspondence structure of SMPL mesh \cite{SMPL:2015}.

The third aspect is how the 3D human generation process is conditioned on different garment layers,  \textit{i.e.,} how to control the generation of human body and garment. 
Existing approaches represent the whole garment layers as one-dimensional latent codes \cite{Fu2022StyleGANHumanAD}, and employ a StyleGAN generator \cite{Karras2019AnalyzingAI} for image synthesis. 
However, these methods perform poorly on layer-wise human generations since the one-dimensional latent representations cannot faithfully encode the 3D structural garment layers. 
To address this, we propose a 3D diffusion model, which generates clothed humans with 3D garment diffusion, and the garment layers can be naturally conditioned in 3D space without losing fidelity.
To fuse the 3D condition into UNet architecture, we adopt a \textbf{3D condition UNet encoder} to project the 3D condition into multi-scale features and hierarchically aggregate them with the diffusion UNet Decoder outputs.
As shown in Fig.~\ref{fig: teaser}, the above design avoids the information loss in processing 3D conditions and successfully learns the 3D condition information through spatially injecting the 3D condition information into the layer-wise 3D human generation process.

We evaluate \nickname\ on two layer-wise 3D human datasets developed from SynBody \cite{synbodyyang2023} and TightCap \cite{chen2021tightcap}. 
Our results show \nickname\ outperforms 3D GAN/Diffusion based models in the layer-wise generation task.
Our contributions are as follows:
\begin{enumerate}
    \item To the best of our knowledge, \nickname{}\ is the first layer-wise 3D human generation approach using diffusion model. It pushes the boundaries of 3D human generation to a more controllable setting and bridges the gap of applying 3D human generation to real-world applications.
    \item \nickname\ proposes 1). a tri-plane shift operation to tackle the challenge of reconstructing fine-grained 3D humans with tri-plane representation; 2). decouples garment styles into different layers (e.g., pants, skirts, shoes) according to the clothed human generation process in the current graphics pipeline and aligns the human body and garments in a canonical space. 3). A 3D condition UNet encoder to hierarchically fuse tri-plane features with the 3D condition in the layer-wise 3D human generation task.
    \item \nickname\ develops two layer-wise 3D human datasets from SynBody (synthetic) and TightCap (real-world), and we will release the two datasets to facilitate more future work in layer-wise 3D human generation. 
\end{enumerate}

%% file: latex/2_related_work.tex
\section{Related Work}
\label{sec:related_work}

\noindent \textbf{Diffusion on 3D.}
While generative diffusion models~\cite{sohl2015deep, song2019generative, song2020denoising, song2020improved,  ho2020denoising} have achieved impressive performance on 2D image generation tasks~\cite{song2020score, saharia2022palette, dhariwal2021diffusion}, extending it to 3D scenes is still a relatively less explored area.
A few recent works build diffusion models based on points, voxels, meshes, or SDF-based 3D representations~\cite{cheng2022sdfusion, hui2022neural, li2022diffusion, luo2021diffusion, zhou20213d, Liu2023MeshDiffusion}. Diffusion models are also proposed to learn neural radiance fields ~\cite{muller2022diffrf}, tri-planes~\cite{wang2022rodin, shue20223d, chen2023single, yang2023learning}, and latent vectors~\cite{nam20223d, ntavelis2023autodecoding}, which enable diverse 3D applications such as novel view synthesis~\cite{watson2022novel}, and 3D reconstruction~\cite{anciukevivcius2022renderdiffusion, liu2023zero}, etc.
Unlike these works, we employ diffusion models in the layer-wise 3D human generation tasks.

\noindent \textbf{Neural Rendering for Humans.} Recent neural rendering methods have made significant progress in generating realistic images of humans by learning from only 2D image collections~\cite{Fruhstuck2022InsetGANFF,Fu2022StyleGANHumanAD, noguchi2022unsupervised, bergman2022generative, hong2022eva3d}. Some existing approaches render human images by neural image translation, \ie they learn the mapping from the renderings of skeletons~\cite{Chan2019edn, SiaroSLS2017, Pumarola_2018_CVPR, KratzHPV2017,zhu2019progressive}, or dense meshes~\cite{Liu2019reenactment, Sarkar2020nrr, Neverova2018, Hu2021ego, Amit21anr,Artur2021stylepeople} to real images by GANs. However, these methods do not reconstruct geometry explicitly and cannot guarantee 3D consistency. For stable view synthesis, some recent papers~\cite{neuralbody,neuralactor,narf,anerf,Weng2022HumanNeRFFR,SHERF,hong2022avatarclip} propose to unify geometry reconstruction with view synthesis by volume rendering~\cite{Kajiya1984RayTV}, which, however, is generally computationally heavy. 
To solve these issues, some 3D-GAN approaches~\cite{Niemeyer2021GIRAFFERS,Gu2021StyleNeRFAS,OrEl2021StyleSDFH3,Hong2021HeadNeRFAR,chan2022efficient} propose a hybrid rendering strategy for efficient geometry-aware rendering based on GANs and volume rendering, which can also be applied to render humans with specific design~\cite{noguchi2022unsupervised,hong2022eva3d,Hu2021HVTRHV,bergman2022generative}. 
However, the aforementioned methods usually assume the garment and body skin belong to the same surface layer, and hence the renderings are not controllable. In contrast, we propose a layer-wise 3D diffusion model which decouples the clothing style into several garment layers and allows users to freely control the generation of the human body and each garment layer.

\noindent \textbf{Human and Garment modeling.} To capture detailed deformations of clothed humans, some recent papers propose to utilize implicit representations for its topological flexibility~\cite{Saito2021SCANimateWS, Mihajlovi2021LEAPLA,Wang2021MetaAvatarLA,Tiwari2021NeuralGIFNG,Jeruzalski2020NASANA,Xiu2022CVPR, He2021ARCHAC}. Most reconstruction works only focus on reconstructing the top surface layer of human and garment modelling is usually included in body modelling~\cite{Saito2021SCANimateWS,Zheng2021PaMIRPM,Saito2019PIFuPI,Saito2020PIFuHDMP}. Contrary to these methods, some other papers propose the idea of multi-layer human modelling, such as DoubleFusion~\cite{Yu2018DoubleFusionRC}, ClothCap~\cite{PonsMoll2017ClothCapS4}, SimuCap~\cite{Yu2019SimulCapS}, and TightCap~\cite{chen2021tightcap}, which separately reconstruct the clothing geometry based on a parametric body template. For these works, 3D scans or depth sensors are required for geometry supervision. Some other approaches propose to generate 3D clothed meshes by learning a data-driven model from 3D datasets~\cite{Ma2019LearningTD, patel20tailornet}. TailorNet~\cite{patel20tailornet} proposes a neural model to predict clothing deformation given pose, shape, and garment style parameters. CAPE~\cite{Ma2019LearningTD} learns a generative model of clothed humans from 3D scans with varying poses and clothing styles. SCARF~\cite{Feng2022CapturingAA} reconstructs human avatars by separately learning body and garment layers from monocular RGB videos. 
Different from these tasks, we focus on learning a 3D generative model from only 2D image collections, and our system allows users to separately control the generation of the human body and garment in a layer-wise manner.

%% file: latex/3_method.tex
\section{Our Approach}
\label{sec:method}

\begin{figure*}[t]
    \centering
    \includegraphics[width=18cm]{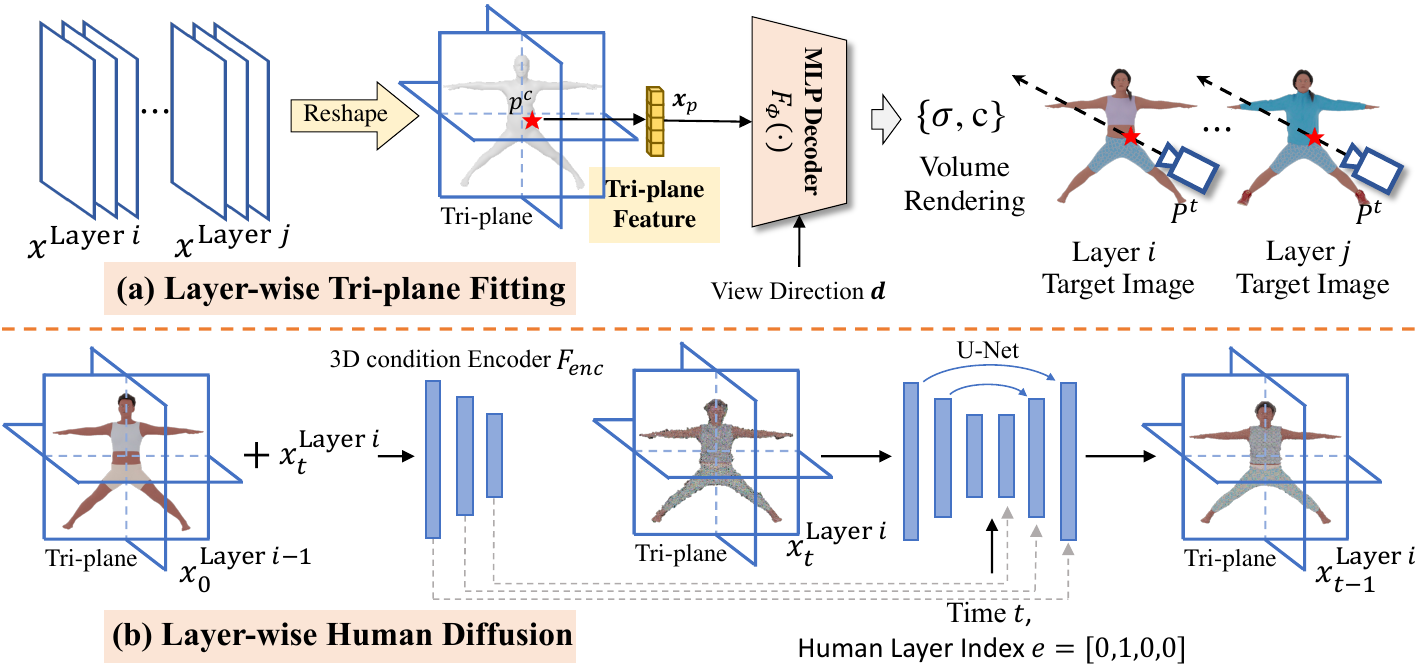}
    \caption{\textbf{Overview of \nickname{}\ Framework.} a) The first stage reconstructs 3D representations, \ie, tri-planes, from multi-view images with a shared decoder. In order to spatially align 3D features, inverse LBS is used to transform humans with different shapes or poses to the canonical spaces. b) In the second stage, we learn layer-wise human diffusion models using the reconstructed tri-planes from the first stage. To facilitate sequential conditional generation, we apply a UNet encoder to hierarchically incorporate tri-plane features  from previous steps with diffused tri-plane features as inputs to the denoising network.}
\label{fig: overview}
\vspace{-5mm}
\end{figure*}

Unlike previous 3D-aware human GAN methods~\cite{chan2022efficient, hong2022eva3d, noguchi2022unsupervised, bergman2022generative, zhang2023avatargen, dong2023ag3d} that learn 3D humans from 2D human image collections, \nickname\ aims to learn layer-wise 3D human generation from 3D representations reconstructed from multi-view images.
Therefore, our pipeline comprises two important steps: \textbf{1)} layer-wise 3D human representation fitting, and \textbf{2)} layer-wise 3D human diffusion model learning.
For the first step, rather than fitting 3D neural renderings independently for each human subject, \nickname\ encodes all human subjects in a canonical space utilizing SMPL inverse Linear Blend Skinning (LBS)~\cite{SMPL:2015} and learns the layer-wise tri-plane representation in this canonical space with a shared MLP NeRF decoder for all subjects.
For the second stage, \nickname\ formulates layer-wise 3D human generation as a sequential 3D conditional generation using diffusion models, where the new layer generation is conditioned on previous 3D representations.
Specifically, \nickname\ first unconditionally generates 3D human representations with minimum clothing.
Then, new layers of clothing are autoregressively generated, where the next layer generation is conditioned on previous 3D representations.

\subsection{Preliminary}
\noindent \textbf{NeRF}~\cite{mildenhall2020nerf} learns an implicit continuous function which takes as input the 3D location $\bm{p}$ and unit direction $\bm{d}$ of each point and predicts the volume density $\bm{\sigma} \in [0, \infty)$ and color value $\bm{c}\in [0,1]^3$, i.e., $F_{\Phi}: (\gamma(\bm{p}), \gamma(\bm{d})) \to (\bm{c}, \bm{\sigma})$, where $F_{\Phi}$ is parameterized by a multi-layer perception (MLP) network, $\gamma$ is the predefined positional embedding applied to $\bm{p}$ and $\bm{d}$.
To render the RGB color of pixels in the target view, rays are cast from the camera origin $\bm{o}$ through the pixel along the unit direction $\bm{d}$. 
Based on the classical volume rendering~\cite{kajiya1984ray}, the expected color $\hat{C}(\bm{r})$ of camera ray $\bm{r}(t) = \bm{o} + t\bm{d}$ is computed as 
\begin{equation}
\label{eqn:volume_rendering}
\begin{aligned}
\hat{C}(\bm{r})=\int_{t_n}^{t_f} T(t) \sigma(\bm{r}(t)) \bm{c}(\bm{r}(t), \bm{d})dt,
\end{aligned}
\end{equation}
where $t_n$ and $t_f$ denote the near and far bounds, $T(t)=\exp(-\int_{t_n}^{t}\sigma(\bm{r}(s))ds)$ denotes the accumulated transmittance along the direction $\bm{d}$ from $t_n$ to $t$.
In practice, the continuous integral is approximated with the quadrature rule~\cite{max1995optical} and reduced to the traditional alpha compositing.

\noindent \textbf{SMPL}~\cite{SMPL:2015} is a parametric human model 
$M(\bm{\beta}, \bm{\theta})$, which defines $\bm{\beta}, \bm{\theta}$ to control body shapes and poses. 
In this work, we apply the Linear Blend Skinning (LBS) algorithm of SMPL to transform points from the canonical space to target/observation spaces. 
Formally, 3D point $\bm{p}^c$ in the canonical space is transformed to an observation space defined by pose $\bm{\theta}$ as $\bm{p}^o = \sum_{k=1}^{K}w_{k}\bm{G_k}(\bm{\theta}, \bm{J})\bm{p}^c$, 
where $K$ is the joint number, $\bm{G_k}(\bm{\theta}, \bm{J})$ is the transformation matrix of joint $k$, $w_{k}$ is the blend weight. 
Similarly, the transformation from target/observation spaces to the canonical space, namely inverse LBS, can be defined with inverted transformation matrices.

\noindent \textbf{Diffusion Model.}~\cite{sohl2015deep} are latent variable models with the form $p_{\theta}(\bm{x}_0):=\int p_{\theta}(\bm{x}_{0:T})d\bm{x}_{1:T}$, where $\bm{x}_1, \dots, \bm{x}_T$ are latent variables with the same dimension as data $\bm{x}_0$. 
The joint distribution $p_{\theta}(\bm{x}_{0:T})$ is defined as the Markov chain with learned Gaussian transitions staring from the standard normal distribution $p(\bm{x}_T)=\mathcal{N}(\bm{x}_T; \bm{0}, \bm{I})$, i.e.,  
\begin{equation}
\label{eqn:reverse_process}
    \begin{aligned}
        & p_{\theta}(\bm{x}_{0:T}):=p_{\theta}(\bm{x}_T) \prod_{t=1}^{T} p_{\theta}(\bm{x}_{t-1}|\bm{x}_t), \\
        & \text{where}\ p_{\theta}(\bm{x}_{t-1}|\bm{x}_t):=\mathcal{N}(\bm{x}_{t-1}; \bm{\mu}_{\theta}(\bm{x}_t, t), \bm{\Sigma}_{\theta}(\bm{x}_t, t)).
    \end{aligned}
\end{equation}

The forward process or diffusion process $q(\bm{x}_{1:T}|\bm{x}_0)$ is defined to be a Markov chain which gradually adds Gaussian noise to data $\bm{x}_0$ according to a variance schedule $\beta_{0}, \dots, \beta_{T}$ defined at different time steps, i.e., 
\begin{equation}
\label{eqn:diffusion_process}
    \begin{aligned}
        & q(\bm{x}_{1:T}|\bm{x}_0):=\prod_{t=1}^{T} q(\bm{x}_{t}|\bm{x}_{t-1}), \\
        & q(\bm{x}_{t}|\bm{x}_{t-1}):=\mathcal{N}(\bm{x}_{t}; \sqrt{\alpha_t}\bm{x}_{t-1}+\beta_{t}\bm{I}).
    \end{aligned}
\end{equation}
where $\alpha_t:=1-\beta_{t}$. With sufficient large $T$ steps, the diffusion process will reach $\bm{x}_T\sim \mathcal{N}(\bm{0}, \bm{I})$. 

Similar to variational auto-encoder~\cite{kingma2019introduction}, diffusion models perform the training based on variational bound on negative log likelihood, i.e.,
\begin{equation}
\label{eqn:elb}
    \begin{aligned}
        \mathbb{E}\left[-\log p_{\theta}(\bm{x}_{0})\right] &\leq\mathbb{E}_{q}\left[-\log \frac{p_{\theta}(\bm{x}_{0:T})}{q(\bm{x}_{1:T}|\bm{x}_0)})\right] \\
        &=\mathbb{E}_{q}\left[-\log p(\bm{x}_T)-\sum_{t\ge 1} \log \frac{p_{\theta}(\bm{x}_{t-1}|\bm{x}_{t})}{q(\bm{x}_{t}|\bm{x}_{t-1})}\right]\\
        &=:\mathcal{L}.
    \end{aligned}
\end{equation}
where during the optimization, the variance $\beta_{t}$ can either be a constant hyper-parameter or a learned parameter, which depends on specific tasks.

\begin{figure*}[t]
    \centering
    \includegraphics[width=18cm]{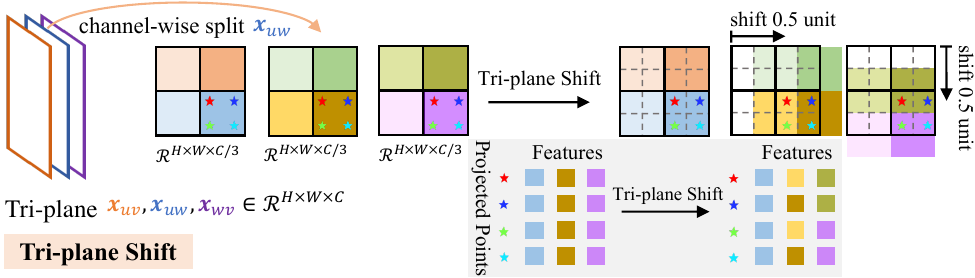}
    \caption{\textbf{Tri-plane Shift.} We split each tri-plane ($\mathcal{R}^{H\times W\times C}$) into three sub-planes ($\mathcal{R}^{H\times W\times C/3}$) and move one sub-plane half unit (0.5/W) to the right, and one sub-plane half unit (0.5/H) downward. With such a design, 3D points projected into the same square area (the light orange, orange, light blue, blue pentagrams) will extract different features and improved reconstruction results can be achieved without enlarging the total number of tri-plane parameters.}
\label{fig: tri-plane shift}
\vspace{-3mm}
\end{figure*}

\subsection{Layer-wise 3D Human Representation Fitting}
To learn 3D human representation for diffusion models, an important step is to find a suitable form of 3D representation satisfying the following requirements.
1) The selected representation should be suitable for generative models like the diffusion model to construct the loss function. 
2) The representation should be memory efficient and have fast convergence speed so that it can be optimized in hours of GPU time.
3) High-fidelity 3D humans can be rendered from this representation. 
4) The layer-wise representation of the same 3D human subject should have similar features for areas with the same cloth or texture. 

Considering these requirements, we select tri-plane representation~\cite{chan2022efficient} for 3D human fitting. 
Tri-plane is an efficient representation which consists of three vertical planes $\bm{x}_{uv}, \bm{x}_{uw}, \bm{x}_{wv} \in \mathcal{R}^{H\times W\times C}$ to form the spatial relationship, where $H, W$ denote spatial resolution and $C$ is the channel number. 
Compared with voxel grid features, tri-plane has much lower memory complexity and is easier to extend to learn high-resolution 3D humans.

During tri-plane fitting as shown in Fig.~\ref{fig: overview}, \nickname\ encodes tri-plane representation in the canonical space to learn neural radiance fields from multi-view images in the target space.
modelling the 3D humans in the canonical space enables generative diffusion models to focus on the learning of human appearance and garments.
The sampled 3D points $\bm{p}^{t}\in \mathcal{R}^{3}$ in the neural radiance field of target space are transformed into $\bm{p}^c$ in the canonical space through SMPL inverse LBS. 
The transformed points $\bm{p}^c$ are projected to three planes $\bm{x}_{uv}, \bm{x}_{uw}, \bm{x}_{wv}$ through orthogonal projection to extract the tri-plane features $\bm{y}_{\bm{p}}\in\mathcal{R}^{3C}$, \ie, 
\begin{equation}
    \begin{aligned}
        \bm{x}_{\bm{p}} = \text{concat}\left[\Pi(\bm{x}_{uv}; \bm{p}^{c}), \Pi(\bm{x}_{uw}; \bm{p}^{c}), \Pi(\bm{x}_{wv}; \bm{p}^{c})\right],
    \end{aligned}
\end{equation}
where $\Pi(\cdot)$ denotes the 3D-to-2D projection operator, $\text{concat}[\cdot]$ denotes the concatenation operator.

The extracted tri-plane features $\bm{x}_{\bm{p}}$ are then fed into NeRF MLP decoder $F_{\Phi}(\cdot)$ along with view direction $\bm{d}$ to predict the corresponding color $\bm{c}(\bm{y}, \bm{d})\in\mathcal{R}^3$ and density $\bm{\sigma}(\bm{y})\in\mathcal{R}^1$, \ie, 
\begin{equation}
    \begin{aligned}
        \bm{c}(\bm{x}, \bm{d}), \bm{\sigma}(\bm{x}) = F_{\Phi}(\gamma(\bm{x}_{\bm{p}}), \gamma(\bm{d})).
    \end{aligned}
\end{equation}
where $\gamma(\cdot)$ is the positional encoding function applied on tri-plane features and view directions. 
The NeRF MLP decoder $F_{\Phi}(\cdot)$ are shared among all human subjects.
After we get the color $\bm{c}(\bm{x}, \bm{d})$ and density $\bm{\sigma}(\bm{x})$ for 3D points $\bm{p}^c$, we perform volume rendering~\cite{kajiya1984ray} to render the predicted image $C(\bm{r})$, where $\bm{r}$ is the corresponding ray containing points $\bm{p}^c$.

To efficiently learn the shared decoder, we randomly sample a portion of layer-wise human subjects to optimize the shared decoder and tri-plane representation.
In the later stage, we fix the decoder and optimize the tri-plane for each subject in a parallel optimization manner. 
To learn similar features for areas with the same cloth or texture of the same human subject, we first fit the tri-plane features $\bm{x}^{\text{Layer}\ i-1}=\text{concat}[\bm{x}_{uv}^{\text{Layer}\ i-1}, \bm{x}_{uw}^{\text{Layer}\ i-1}, \bm{x}_{wv}^{\text{Layer}\ i-1}]$ for the $(i-1)$th human layer and then use this as the initialization for next-layer tri-plane $\bm{x}^{\text{Layer}\ i}=\text{concat}[\bm{x}_{uv}^{\text{Layer}\ i}, \bm{x}_{uw}^{\text{Layer }\ i}, \bm{x}_{wv}^{\text{Layer}\ i}]$ fitting. 
During our experiments, we find such an optimization scheme improves both the reconstruction and the later layer-wise 3D human generation results.

In the tri-plane representation, the resolution size $H\times W$ determines the details of final reconstruction results as 3D points projected to the same square area of the plane share the same features. 
To further improve the reconstruction quality, we propose a \textbf{tri-plane shift} operation as shown in Fig~\ref{fig: tri-plane shift}.
Specifically, we split each tri-plane into 3 sub-planes (each with size $\mathcal{R}^{H\times W\times C/3}$) and move one sub-plane half unit ($0.5/W$) to the right, and one sub-plane half unit ($0.5/H$) downward.
With such design, points projected to the same square area in each plane will extract different features, and the extracted features can therefore be used to render more fine-grained results.
Later ablation study shows that our proposed \textbf{tri-plane shift} improves the 3D human reconstruction performance without enlarging the total number of tri-plane parameters.

In the optimization stage, four loss functions are used to optimize tri-plane features, \ie,\\
\noindent \textbf{Photometric Loss.} 
Given the ground truth target image $C(\mathbf{r})$  and predicted image $\hat{C}(\mathbf{r})$, we apply the photometric loss as follows:
\begin{equation}
\label{loss: color}
\begin{aligned}
\mathcal{L}_{color} = \frac{1}{|\mathcal{R}|}\sum_{\mathbf{r}\in \mathcal{R}} ||\hat{C}(\mathbf{r}) - C(\mathbf{r})||_2^2,
\end{aligned}
\end{equation}
where $\mathcal{R}$ denotes the set of rays, and $|\mathcal{R}|$ is the number of rays.

\noindent \textbf{Mask Loss.} 
We also leverage the human region masks for NeRF optimization. The mask loss is defined as:
\begin{equation}
\label{loss: mask}
\begin{aligned}
\mathcal{L}_{mask} = \frac{1}{|\mathcal{R}|}\sum_{\mathbf{r}\in \mathcal{R}} ||\hat{M}(\mathbf{r}) - M(\mathbf{r})||_2^2,
\end{aligned}
\end{equation}
where $\hat{M}(\mathbf{r})$ is the accumulated volume density and $M(\mathbf{r})$ is the ground truth binary mask label.

\noindent \textbf{Total Variation (TV) loss.} 
We encourage the tri-plane features to be smooth. The total variation (TV) loss regularity is defined as:
\begin{equation}
\label{loss: tv loss regularity}
\begin{aligned}
\mathcal{L}_{tv} & = \frac{1}{N}\sum_{n=1}^{N}||\bm{x}_{uv}[n+1] - \bm{x}_{uv}[n]||_{1} \\
& + \frac{1}{N}\sum_{n=1}^{N}||\bm{x}_{uw}[n+1] - \bm{x}_{uw}[n]||_{1} \\
& + \frac{1}{N}\sum_{n=1}^{N}||\bm{x}_{wv}[n+1] - \bm{x}_{wv}[n]||_{1},
\end{aligned}
\end{equation}
where $n$ denotes the feature index in the width or height direction, $N$ is the number of features in each tri-plane.   

\noindent \textbf{Feature sparsity Regularity.} 
We also encourage the tri-plane features to be sparse so that only points near the 3D human surface have meaningful feature values. The feature sparsity regularity is defined as:
\begin{equation}
\label{loss: feature sparsity regularity}
\begin{aligned}
\mathcal{L}_{feature} & = \frac{1}{N} \sum_{\bm{x}_{uv}} || \bm{x}_{uv} - 0||_1 + \frac{1}{N} \sum_{\bm{x}_{uw}}|| \bm{x}_{uw} - 0||_1 \\
& + \frac{1}{N} \sum_{\bm{x}_{wv}}|| \bm{x}_{wv} - 0||_1.
\end{aligned}
\end{equation}

\noindent In summary, the overall loss function contains three components, \ie, 
\begin{equation}
\label{loss: overall}
\begin{aligned}
\mathcal{L} = \mathcal{L}_{color} + \lambda_{1}\mathcal{L}_{mask} + \lambda_{2}\mathcal{L}_{tv} + \lambda_{3}\mathcal{L}_{feature},
\end{aligned}
\end{equation}
where $\lambda$'s are the loss weights. 
Empirically, we set $\lambda_1=0.1$, $\lambda_2 = 0.01$, $\lambda_3 = 0.0005$ to ensure the same magnitude for each loss term.

\subsection{Layer-wise 3D Human Diffusion}
After fitting the layer-wise tri-plane features, \nickname\ learns layer-wise 3D human generation as the sequential 3D conditional generation, which generates the 3D human tri-plane presentation (\eg, $\bm{x}^{\text{Layer}\ i}$) conditioned on tri-plane features (\eg, $\bm{x}^{\text{Layer}\ i-1}$) generated from the previous human layer.

\noindent \textbf{Layer-wise Diffusion Process.} The diffusion process is defined as a Markov process that adds noise to the sampled data according to the noise schedule at time step $t$. In the forward process or diffusion process (as shown in Eqn.~\ref{eqn:diffusion_process}) of layer-wise 3D human diffusion, diffusion model starts from taking $\bm{x}_0^{\text{Layer}\ i}$ from the 3D human tri-plane distribution, iteratively samples $\bm{x}_t^{\text{Layer}\ i}$ given $\bm{x}_{t-1}^{\text{Layer}\ i}$, and stops until $\bm{x}_T^{\text{Layer}\ i}$ completely reaches random noise.
For arbitrary time step $t$, we can derive the distribution $q(\bm{x}_{t}^{\text{Layer}\ i}|\bm{x}_0^{\text{Layer}\ i})$ utilizing the Gaussian distribution property:
\begin{equation}
    \begin{aligned}
        q(\bm{x}_{t}^{\text{Layer}\ i}|\bm{x}_0^{\text{Layer}\ i}):=\mathcal{N}(\bm{x}_{t}^{\text{Layer}\ i}; \sqrt{\bar{\alpha}_t}\bm{x}_{0}^{\text{Layer}\ i}+(1-\bar{\alpha}_t)\bm{I}),
    \end{aligned}
\end{equation}
where $\bar{\alpha}_t:=\prod_{i=1}^{t}\alpha_i$. This equation let us quickly get the diffused samples at arbitrary time steps.

\noindent \textbf{Layer-wise Denoising Process.} is also defined as a Markov chain with the same time steps $\{0, \dots, T\}$. 
The denoising process starts by sampling $\bm{x}_T^{\text{Layer}\ i}$ from the normal distribution $\mathcal{N}(\bm{0}, \bm{I})$ and generates the $\bm{x}_{t-1}^{\text{Layer}\ i}$ from $\bm{x}_{t}^{\text{Layer}\ i}$ and previous human layer output $\bm{x}_{0}^{\text{Layer}\ i-1}$. 
The state transition probability $p_{\theta}(\bm{x}_t^{\text{Layer}\ i}|\bm{x}_{t-1}^{\text{Layer}\ i}, \bm{x}_{0}^{\text{Layer}\ i-1})$ is modelled by the U-Net~\cite{ronneberger2015u} with parameters $\theta$, which has the following form
\begin{equation}
    \begin{aligned}
        & p_{\theta}(\bm{x}_{t-1}^{\text{Layer}\ i}|\bm{x}_t^{\text{Layer}\ i}, \bm{x}_{0}^{\text{Layer}\ i-1}):=\mathcal{N}(\bm{x}_{t-1}^{\text{Layer}\ i}; \\
        & \bm{\mu}_{\theta}(\bm{x}_t^{\text{Layer}\ i}, \bm{x}_{0}^{\text{Layer}\ i-1}, t), \bm{\Sigma}_{\theta}(\bm{x}_t^{\text{Layer}\ i}, \bm{x}_{0}^{\text{Layer}\ i-1}, t)),
    \end{aligned}
\end{equation}
where $\bm{\mu}_{\theta}(\bm{x}_t^{\text{Layer}\ i}, \bm{x}_{0}^{\text{Layer}\ i-1}, t)=\frac{1}{\sqrt{\alpha_t}}\bm{x}_t - \frac{1-\alpha_t}{\sqrt{1-\bar{\alpha}_t}\sqrt{\alpha_t}}\hat{\bm{\epsilon}}_{\theta}(\bm{x}_t^{\text{Layer}\ i}, \bm{x}_{0}^{\text{Layer}\ i-1}, t)$, $\hat{\bm{\epsilon}}_{\theta}(\bm{x}_t^{\text{Layer}\ i}, \bm{x}_{0}^{\text{Layer}\ i-1}, t)$ is predicted from U-Net with noised sample $\bm{x}_t^{\text{Layer}\ i}$ as input, previous human layer generation output $\bm{x}_{0}^{\text{Layer}\ i-1}$ and time step $t$ as condition.
The variance term can be derived in a similar form. 

It is non-trivial to directly combine the 3D condition $\bm{x}_{0}^{\text{Layer}\ i-1}$ into U-Net using adaptive group normalization (AdaGN)~\cite{xu2019understanding} or cross-attention, as the above two condition methods map 3D conditions into one-dimensional latent codes. 
The compressed latent codes lose the spatial 3D information crucial in the conditional generation process because we need to preserve the identity or garment color from previous layers. 
Drawing inspiration from ControlNet~\cite{zhang2023adding} which shows remarkable control ability over 2D conditions, we adopt a similar way to process 3D conditions. 
Specifically, we use a UNet encoder to extract multi-level 3D condition feature information from the 3D condition and hierarchically fuse them to guide the learning of diffusion UNet decoder.
As shown in Fig.~\ref{fig: overview}, we first pre-process the 3D condition $\bm{x}_{0}^{\text{Layer}\ i-1}$ by adding the diffused sample $\bm{x}_{t}^{\text{Layer}\ i}$ and encode it with the UNet encoder $F_{enc}(\cdot)$ to extract 3D condition at different spatial levels.
Then we fuse the multi-level 3D condition features to control the outputs of each level of diffusion UNet decoder blocks.
With such a design, we hierarchically incorporate 3D conditions into diffusion models, leveraging coarse and fine 3D condition information to help preserve the corresponding human body and garment information from the previous layer.


\noindent \textbf{Layer-wise Diffusion Training Objective.}
Taking variance reduction approach into Eqn.~\ref{eqn:elb} and following~\cite{ho2020denoising}, the training objective can be derived as  
\begin{equation}
    \begin{aligned}
        &\mathcal{L}_{simple}:=\mathbb{E}_{t, \bm{x}_t^{\text{Layer}\ i}, \epsilon}\left[||\hat{\bm{\epsilon}}_{\theta}(\bm{x}_t^{\text{Layer}\ i}, \bm{x}_{0}^{\text{Layer}\ i-1}, t) - \bm{\epsilon}||_2^2\right]=\\
        & \mathbb{E}_{t, \bm{x}_t^{\text{Layer}\ i}, \epsilon}\left[||\hat{\bm{\epsilon}}_{\theta}(\bm{x}_t^{\text{Layer}\ i}, F_{enc}(\bm{x}_t^{\text{Layer}\ i}+\bm{x}_{0}^{\text{Layer}\ i-1}), t) - \bm{\epsilon}||_2^2\right],
    \end{aligned}
\end{equation}
where noise $\bm{\epsilon}$ follows a standard normal distribution. For the human template $\bm{x}_{0}^{\text{Layer}\ 0}$ (first human layer) generation, we set the 3D condition as tensors with zero elements. Considering the compactness and scalability, we model the sequential 3D conditional generation in one U-Net. To facilitate that, we use an additional one-hot vector $e$ to denote which layer is generated, \ie,       
\begin{equation}
    \begin{aligned}
        &\mathcal{L}_{simple}:=\mathbb{E}_{t, \bm{x}_t^{\text{Layer}\ i}, \epsilon}\left[||\hat{\bm{\epsilon}}_{\theta}(\bm{x}_t^{\text{Layer}\ i}, \bm{x}_{0}^{\text{Layer}\ i-1}, t, e) - \bm{\epsilon}||_2^2\right]=\\
        &\mathbb{E}_{t, \bm{x}_t^{\text{Layer}\ i}, \epsilon}\left[||\hat{\bm{\epsilon}}_{\theta}(\bm{x}_t^{\text{Layer}\ i}, F_{enc}(\bm{x}_t^{\text{Layer}\ i}+\bm{x}_{0}^{\text{Layer}\ i-1}), t, e) - \bm{\epsilon}||_2^2\right].
    \end{aligned}
\end{equation}
For example, in a four-layer generation task, $e=[0,1,0,0]$ corresponds to human layer 1 generation.

%% file: latex/4_experiment.tex
\section{Experiment}
\label{sec:exp_section}

\begin{figure*}[t]
    \centering
    \includegraphics[width=.92\linewidth]{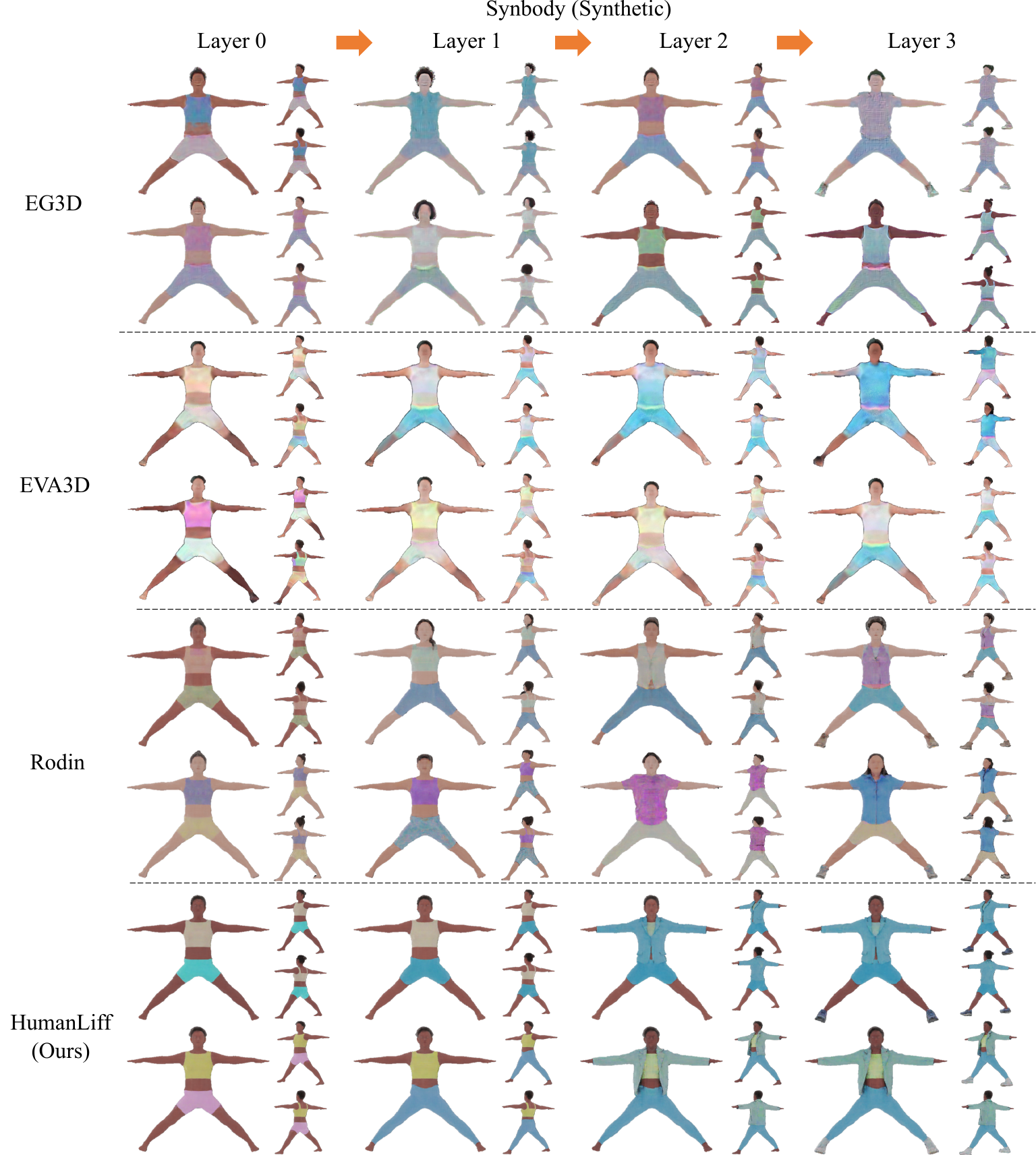}
    \setlength{\abovecaptionskip}{0.1cm}
    \caption{Layer-wise 3D human generation results produced by EG3D, EVA3D, Rodin and our \textbf{\nickname{}} on the SynBody dataset. Each row (from Layer 0 to Layer 3) contains layer-wise 3D human generation results for the \textbf{same human subject}.} 
\label{fig: synbody_visual_result}
\vspace{-3mm}
\end{figure*}

\begin{figure*}[t]
    \centering
    \includegraphics[width=.92\linewidth]{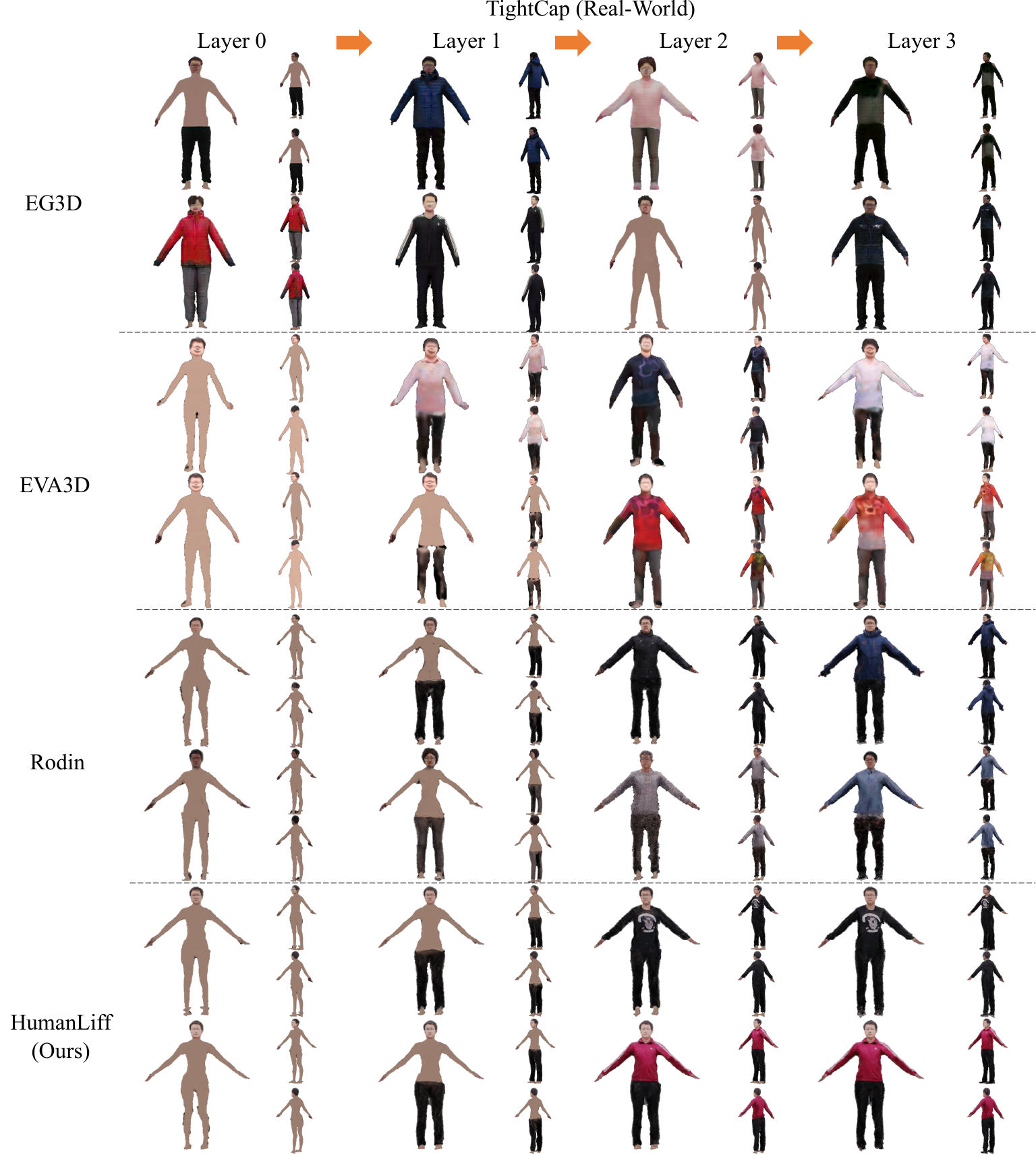}
    \setlength{\abovecaptionskip}{0.1cm}
    \caption{Layer-wise 3D human generation results produced by EG3D, EVA3D, Rodin and our \textbf{\nickname{}} on the TightCap dataset. Each row (from Layer 0 to Layer 3) contains layer-wise 3D human generation results for the \textbf{same human subject}.} 
\label{fig: tightcap_visual_result}
\vspace{-4mm}
\end{figure*}

\begin{table*}[t]
\setlength{\abovecaptionskip}{0cm}
\caption{Performance (FID and L-PSNR) comparison among EG3D, EVA3D, Rodin and our \nickname{} model on layer-wise SynBody and TightCap datasets.}
\centering
\label{tab: main_redult}
\begin{tabular}{l|cccccccc|c}
\toprule
\multirow{3}*{Method} & \multicolumn{9}{c}{\textit{SynBody} (synthetic)}  \\
\cline{2-10}
 & \multicolumn{2}{c}{Layer 0} & \multicolumn{2}{c}{Layer 1} & \multicolumn{2}{c}{Layer 2} & \multicolumn{2}{c}{Layer 3} & Unconditional\\
\cline{2-10}  
~ & FID$\downarrow$ & L-PSNR$\uparrow$ & FID$\downarrow$ & L-PSNR$\uparrow$ & FID$\downarrow$ & L-PSNR$\uparrow$ & FID$\downarrow$ & L-PSNR$\uparrow$ & FID$\downarrow$\\
\hline\hline
EG3D~\cite{chan2022efficient} & 99.20 & - & 99.01 & $<$20 & 106.31 & $<$20 &  110.74 & $<$20 & 70.00 \\
EVA3D~\cite{hong2022eva3d} & 104.36 & - & 119.18 & $<$20 & 121.66 & $<$20 & 123.58 & $<$20 & 104.73\\
Rodin~\cite{wang2022rodin} & 63.49 & - & 55.12 & 19.54 & 38.02 & 20.27 & 41.91 & 21.26 & 23.29\\
\textbf{\nickname{} (Ours)} & \textbf{22.28} & - & \textbf{25.09} & \textbf{28.14} & \textbf{24.95} & \textbf{28.44} & \textbf{25.20} & \textbf{29.87} & \textbf{21.40}\\
\midrule
\multirow{3}*{Method} & \multicolumn{9}{c}{\textit{TightCap} (real-world)}  \\
\cline{2-10}
 & \multicolumn{2}{c}{Layer 0} & \multicolumn{2}{c}{Layer 1} & \multicolumn{2}{c}{Layer 2} & \multicolumn{2}{c}{Layer 3} & Unconditional\\
\cline{2-10}  
~ & FID$\downarrow$ & L-PSNR$\uparrow$ & FID$\downarrow$ & L-PSNR$\uparrow$ & FID$\downarrow$ & L-PSNR$\uparrow$ & FID$\downarrow$ & L-PSNR$\uparrow$ & FID$\downarrow$\\
\hline\hline
EG3D~\cite{chan2022efficient} & 112.55 & - & 91.40 & $<$20 & 93.12 & $<$20 & 89.04 & $<$20 & 75.42 \\
EVA3D~\cite{hong2022eva3d} & 64.22 & - & 79.45 & $<$20 & 121.64 & $<$20 & 118.50  & $<$20 & 61.58 \\
Rodin~\cite{wang2022rodin} & 53.92 & - & 81.24 & 21.48 & 132.40 &  17.64 & 92.03 & 15.23 & 56.57\\
\textbf{\nickname{} (Ours)} & \textbf{50.26} & - & \textbf{64.63} & \textbf{28.69} & \textbf{77.19} & \textbf{28.13} & \textbf{65.80} & \textbf{28.90} & \textbf{54.39} \\
\bottomrule
\end{tabular}
\vspace{-3mm}
\end{table*}

\subsection{Experimental Setup}
\noindent \textbf{Datasets.}
We evaluate the performance of \nickname\ on two layer-wise 3d human datasets developed from SynBody (synthetic)~\cite{yang2023synbody} and TightCap (real-world)~\cite{chen2021tightcap}.
For the layer-wise SynBody dataset, we obtain 1000 subjects with a random sampling of human body textures, pants, shirts and shoes.
As shown in Fig.~\ref{fig: synbody_visual_result}, each subject has four layers (\ie, body, pants, shirts, shoes). For each subject and each layer, we render 185 multi-view images under spherical camera poses using Blender engine. 
These multi-view images are used for tri-plane fitting and GAN-based baseline training.
For the layer-wise TightCap dataset, we obtain 107 human subjects from the original dataset.
Similar to the layer-wise SynBody dataset, we render 185 multi-view images for each subject and each layer.
For both layer-wise datasets, we apply SMPLify
~\cite{bogo2016keep} to estimate SMPL parameters based on 3D key-points (2D key-points triangulation) and 3D mesh. 

\noindent \textbf{Comparison Methods.}
To the best of our knowledge, we are the first to study the setting of layer-wise 3D human generation using diffusion models.
For performance comparison, we adopt two state-of-the-art GAN and diffusion based 3D generative models, \ie, EG3D~\cite{chan2022efficient}, EVA3D~\cite{hong2022eva3d} and Rodin~\cite{wang2022rodin}.
EG3D and EVA3D are designed for unconditional 3D generation. 
It is not trivial for EG3D and EVA3D to perform layer-wise generation. 
Therefore, we apply two modifications to the original EG3D and EVA3D for the layer-wise setting. 
1) Similar to \nickname, we encode the layer index one-hot vector $e$ into the condition vector $c$.
To further improve the discrimination ability of the discriminator, we also augment the real image with the wrong layer index one-hot vector as the fake sample. 
2) To inject 3D conditions into EG3D, we project the tri-plane features from the previous layer into latent codes and use them as the condition.
For the baseline Rodin, we follow the original paper to implement the roll-out, 3D-aware convolution operations and project the tri-plane condition into one-dimensional latent vectors, which serve as conditions in an AdaGN form. 
The human layer index vector $e$ is also added into Rodin as the layer index indicator.

\noindent \textbf{Implementation Details.}
For tri-plane fitting, we set the dimension of each tri-plane to be $256\times 256\times 9$. 
By incorporating the tri-plane shift, each tri-plane is split into three sub-planes, each with dimension $256\times 256\times 3$.
We first sample 400 subjects from layer-wise SynBody to learn the shared 4-layer MLP decoder and then fit each tri-plane in a parallel manner with shared decoder parameters fixed.
For efficient ray points sampling, we sample 128 coarse points and use importance sampling to further sample 128 fine points based on the shared MLP decoder. 
During optimization, we use Adam optimizer~\cite{kingma2014adam} and set the learning rate for MLP decoder parameters to be $5\times10^{-3}$ and tri-plane parameters to be $1\times10^{-1}$.
The above fitting pipeline is also used for layer-wise TightCap tri-plane fitting.

For layer-wise diffusion model, we adopt 2D-UNet model introduced in~\cite{ho2020denoising, nichol2021improved} to predict $\hat{\epsilon}_{\theta}$.
The input and output channel number of the 2D-UNet model is set to 27.
We set the diffusion time steps to be 1000 and the middle channel number of 2D-UNet to be 192.
Linear noise schedule with the default setting in~\cite{nichol2021improved} is applied, and the learning rate is set as $5\times10^{-5}$. 
For the 3D condition UNet encoder, we adopt similar architecture as the encoder in the layer-wise diffusion model and apply a zero convolution (1x1 convolution with both weight and bias initialized as zeros) to project the 3D condition features at different levels before incorporating them into the layer-wise diffusion model.
During the inference stage of the diffusion model, we set the time steps to 250 to enable efficient sampling.  
All our experiments are conducted on NVIDIA Tesla V100 GPU cards.

\noindent \textbf{Evaluation Metrics.}
To quantitatively evaluate the quality of generated samples, we adopt Fréchet Inception Distance (FID)~\cite{heusel2017gans} on renderings of generated tri-planes.
We also measure how well the corresponding layer condition information is preserved during layer-wise generation.
Specifically, we mask out newly generated parts and report layered peak signal-to-noise ratio (L-PSNR)~\cite{sara2019image} on the rest of the generated renderings.

\subsection{Quantitative Results}
As shown in Tab.~\ref{tab: main_redult}, \nickname\ significantly outperforms EG3D and EVA3D in both FID and L-PSNR in both SynBody (synthetic) and TightCap (real-world) datasets.
EG3D and EVA3D, mainly designed for unconditional 3D generation, fails to produce high-quality results in the layer-wise 3D human generation. 
There are two main reasons behind this.
1) EG3D and EVA3D can not effectively learn from the human layer index condition to generate correct layer-specific results, even if we augment the real image with the wrong layer index condition as fake in the training.  
2) EG3D and EVA3D can not faithfully recover the 3D structural information from the projected tri-plane latent code condition, leading to the almost unconditional generation.
Compared with the diffusion-based Rodin model, \nickname{} achieves lower FID scores and higher L-PSNR in the layer-wise human generation task.
In our experiments, we observe Rodin does not generate correct samples corresponding to the 3D condition. 
One possible reason is that encoding 3D layered conditions in one-dimensional latent codes using the AdaGN form loses structural 3D information.
Due to this, Rodin performs much worse in L-PSNR and FID, while our \nickname can learn the corresponding 3D condition information using the proposed 3D condition UNet encoder.
In comparison with GAN-based EG3D, EVA3D and diffusion-based Rodin, our \nickname\ successfully learn a sequence of diffusion-based conditional generation in one unified diffusion model.
To further verify the correctness of our implemented baseline methods, we also show the comparison results of unconditional 3D human generation tasks in both SynBody (synthetic) and TightCap (real-world) datasets.
As shown in Tab.~\ref{tab: main_redult}, GAN-based EG3D and EVA3D, diffusion-based Rodin methods all achieve much lower FID when compared with layer-wise 3D human generation task.
The above suggests the layer-wise 3D human generation is a more challenging task than unconditional 3D human generation.

\begin{table*}[t]
\setlength{\abovecaptionskip}{0cm}
\caption{Ablating 3D conditioning methods, \ie AdaGN, cross-attention and 3D UNet Encoder, on the layer-wise SynBody dataset.}
\centering
\label{tab: ablation_condition_redult}
\begin{tabular}{l|cccccccc}
\toprule
\multirow{3}*{Method} & \multicolumn{8}{c}{\textit{SynBody} (synthetic)}  \\
\cline{2-9}
 & \multicolumn{2}{c}{Layer 0} & \multicolumn{2}{c}{Layer 1} & \multicolumn{2}{c}{Layer 2} & \multicolumn{2}{c}{Layer 3} \\
\cline{2-9}  
~ & FID$\downarrow$ & L-PSNR$\uparrow$ & FID$\downarrow$ & L-PSNR$\uparrow$ & FID$\downarrow$ & L-PSNR$\uparrow$ & FID$\downarrow$ & L-PSNR$\uparrow$\\
\hline\hline
AdaGN & 23.38 & - & 28.37 & 20.96 & 27.64 & 22.71 & 26.83 & 22.93 \\
Cross-attention & 26.62 & - & 28.23 & 27.18 & 27.96 & 28.25 & 27.94 & 29.62 \\
UNet Encoder (Ours) & \textbf{22.28} & - & \textbf{25.09} & \textbf{28.14} & \textbf{24.95} & \textbf{28.44} & \textbf{25.20} & \textbf{29.87}\\
\bottomrule
\end{tabular}
\end{table*}

\begin{figure*}[t]
    \centering
    \includegraphics[width=18cm]{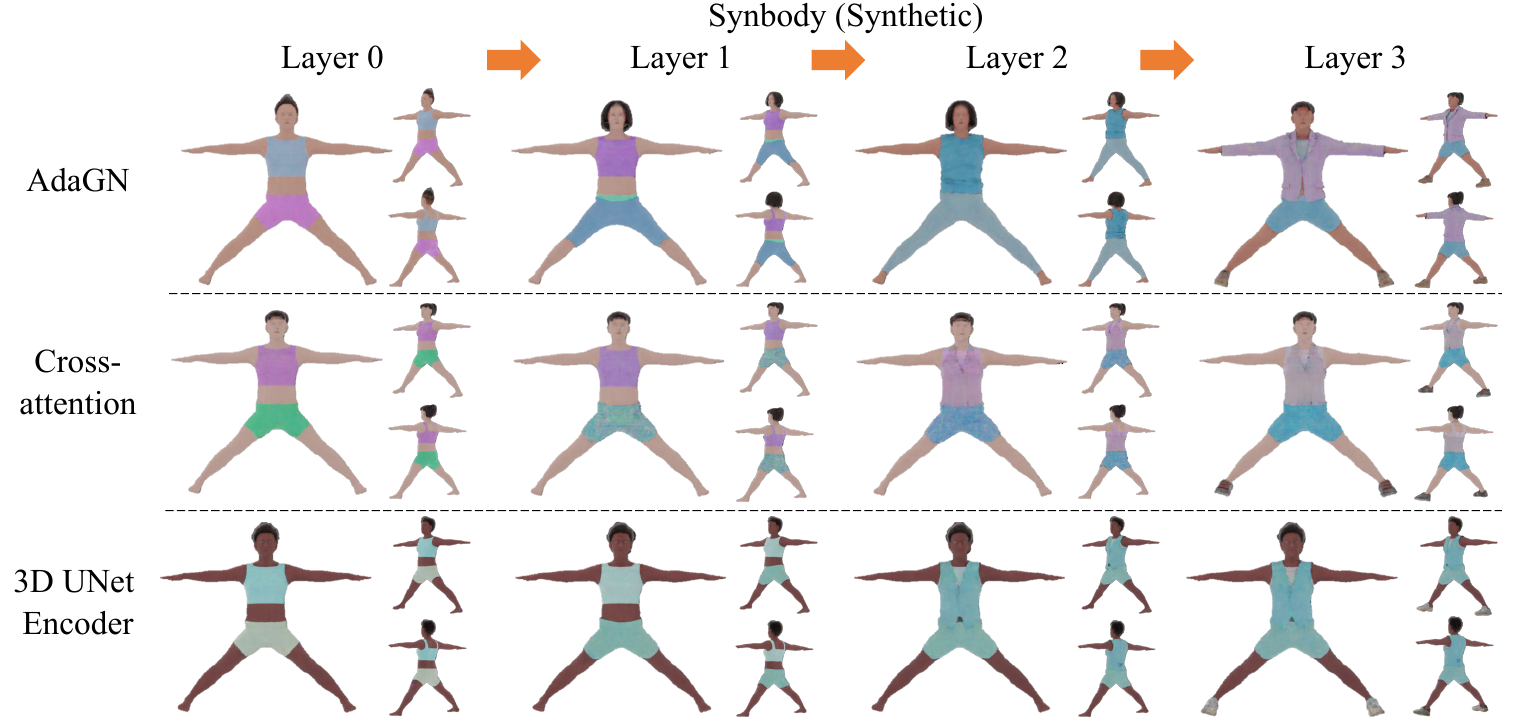}
    \setlength{\abovecaptionskip}{-0cm}
    \caption{Qualitative results of ablating 3D conditioning methods in the layer-wise 3D human generation on the SynBody dataset. Each row (from Layer 0 to Layer 3) contains the layer-wise 3D human generation results for the \textbf{same human subject}.} 
\label{fig: ablation_condition_result}
\vspace{-3mm}
\end{figure*}

\subsection{Qualitative Results}
We show rendered images of our \nickname\ and above three baseline methods in Fig.~\ref{fig: synbody_visual_result} and Fig.~\ref{fig: tightcap_visual_result}.
EG3D and EVA3D fail to produce 3D human details, \eg, face and garment details.
EG3D and EVA3D also fail to learn the human layer index and 3D human layer conditions. 
Rodin successfully learns the human layer index condition but fails to preserve identities during generation.
As shown in Fig.~\ref{fig: synbody_visual_result} and Fig.~\ref{fig: tightcap_visual_result}, different layered outputs produced by Rodin show little correlation, and the quality of Rodin outputs are worse than our \nickname.
Thanks to our fusion of tri-plane features with layered conditions, \nickname\ successfully generates 3D humans that align with layered conditions.
For example, except for the newly generated part, the generated Layer $i$ human keeps almost the same information when compared with Layer $i-1$ human condition.

\subsection{Ablation Study}
\noindent \textbf{Ablating tri-plane shift.}
To validate the effectiveness of the proposed tri-plane shift operation in layer-wise 3D human reconstruction, we show the comparison results w/o and w/ tri-plane shift operation on the SynBody (synthetic) dataset in Tab.~\ref{tab: ablation_triplane_redult} and Fig.~\ref{fig: ablation_tri_plane_redult}.
As shown in Tab.~\ref{tab: ablation_triplane_redult}, by using the same number of tri-plane parameters, tri-plane representation equipped with tri-plane shift improves the PSNR, SSIM and LPIPs metrics, leading to better reconstruction results.
For example, the tri-plane shift can help reconstruct more details in the eye area, as shown in Fig.~\ref{fig: ablation_tri_plane_redult}.

\begin{table}[h]
\caption{Ablation study on 3D conditioning methods with AdaGN and cross-attention on the layer-wise SynBody dataset.}
\centering
\label{tab: ablation_triplane_redult}
\setlength{\abovecaptionskip}{-0.2cm}
\begin{tabular}{l|ccccc}
\toprule
\multirow{2}*{Method} & \multicolumn{3}{c}{\textit{SynBody} (synthetic)} \\
\cline{2-4}
~ & PSNR$\uparrow$ & SSIM$\uparrow$ & LPIPs$\downarrow$ \\
\midrule
w/o tri-plane shift & 31.35 & 0.97 & 0.05 \\
w/ tri-plane shift & \textbf{31.97} & \textbf{0.98} & \textbf{0.04} & \\
\bottomrule
\end{tabular}
\end{table}

\begin{figure}[h]
    \centering
    \includegraphics[width=.98\linewidth]{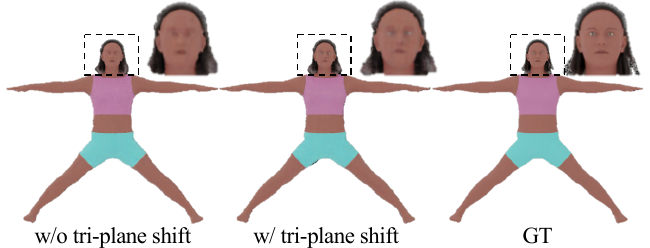}
    \setlength{\abovecaptionskip}{0.1cm}
    \caption{Qualitative results of ablation study on layer-wise human reconstruction results w/o and w/ tri-plane shift on the SynBody dataset.} 
\label{fig: ablation_tri_plane_redult}
\vspace{-3mm}
\end{figure}

\noindent \textbf{Ablating 3D condition methods.}
To validate the effectiveness of the proposed fusion of tri-plane features and 3D layer condition, we compare it with two commonly used 3D conditioning methods in 2D and 3D generative models, \ie, AdaGN and cross-attention, on the SynBody (synthetic) dataset in Tab.~\ref{tab: ablation_condition_redult} and Fig.~\ref{fig: ablation_condition_result}. 
AdaGN and cross-attention require the condition to be one-dimensional latent codes, so we project the 3D layered condition into one-dimensional latent codes using convolution operations.
As shown in Tab.~\ref{tab: ablation_condition_redult}, AdaGN produces much worse FID and L-PSNR when compared with our proposed 3D condition UNet Encoder module.
This indicates diffusion models can not effectively learn 3D condition information from the AdaGN condition method. 
As shown in Fig.~\ref{fig: ablation_condition_result}, Cross-attention can learn the 3D condition information to some extent by injecting the projected one-dimensional latent codes into the cross-attention module. 
However, due to the information loss in projected one-dimensional latent codes, cross-attention fails to generate 3D human with the same garment color as the 3D condition. 
Compared with AdaGN and cross-attention, our proposed 3D condition UNet encoder can produce almost the same face and garment details from multi-level 3D condition features in the layer-wise 3D human generation.

%% file: latex/5_discussion.tex
\section{Discussion and Conclusion}
To conclude, we propose \nickname, the first layer-wise 3D human generative model with a unified diffusion process.
Specifically, \nickname\ encodes the 3D human with tri-plane representation in a canonical space and learns to generate 3D humans in a layer-wise manner.
To render high-fidelity 3D humans, \nickname\ proposes a tri-plane shift operation to improve 3D human reconstruction details. 
To further enhance the controllability of layer-wise 3D generation with the layered condition, \nickname\ proposes a 3D condition UNet encoder to hierarchical fuse multi-level 3D condition features with diffusion UNet decoder outputs to facilitate the learning of the 3D Diffusion model.
Extensive experiments on SynBody (synthetic) and TightCap (real-world) datasets validate that \nickname\ achieves state-of-the-art performance in the layer-wise 3D human generation task.

\vspace{1.75mm}
\noindent \textbf{Limitations:}
1) Reconstructing 3D human using the tri-plane representation can produce reasonable rendering results, but reconstructing high-fidelity 3D human from multi-view images still remains a challenging problem to tackle.
One promising direction is to propose a new 3D human representation that can produce both high-fidelity 3D human reconstruction and generation results.
2) There still exists some artifacts in the generation renderings.
One possible reason is that there exists gap between diffusion training and inference in the layer-wise generation task as the 3D condition generated from previous human layer diffusion differs from the 3D ground truth condition used in the training.
One potential direction is involving generated 3D tri-plane (condition) in the diffusion training to mitigate the gap. 
3) Our current \nickname\ did not consider the interaction with other modalities, like text or speech. 
To involve such conditions into the layer-wise 3D human generation remains a future work to explore. 